%% file: segstereo.tex
\documentclass[runningheads]{llncs}
\usepackage{graphicx}
\usepackage{amsmath,amssymb} 
\usepackage{color}

\usepackage{stfloats}
\usepackage{tablefootnote}
\usepackage{algorithm}
\usepackage{algorithmic}
\usepackage{booktabs}
\usepackage{array}
\usepackage{multirow}
\usepackage{caption}
\usepackage{subcaption}
\usepackage[pagebackref=true,breaklinks=true,letterpaper=true,colorlinks,bookmarks=false]{hyperref}

\newcommand{\etal}{\textit{et al}.}

\newcommand\blfootnote[1]{
  \begingroup
  \renewcommand\thefootnote{}\footnote{#1}
  \addtocounter{footnote}{-1}
  \endgroup
}

\begin{document}
\title{SegStereo: Exploiting Semantic Information for Disparity Estimation} 

\titlerunning{SegStereo}
\author{
{Guorun Yang}${^{1\star}}$ \and
{Hengshuang Zhao}${^{2\star}}$ \and
{Jianping Shi}${^{3}}$ \and \\
{Zhidong Deng}${^{1}}$ \and
{Jiaya Jia}${^{2,4}}$
}

\institute{${^1}${Department of Computer Science, Tsinghua University$^\dagger$} \\
${^2}${The Chinese University of Hong Kong} \\
${^3}${SenseTime Research}, ${^4}${Tencent YouTu Lab} \\
\email{ \{ygr13@mails,michael@mail\}.tsinghua.edu.cn, \\
\{hszhao,leojia\}@cse.cuhk.edu.hk, shijianping@sensetime.com}
}

\authorrunning{G. Yang, H. Zhao, J. Shi, Z. Deng and J. Jia}

\maketitle              

\blfootnote{$^{\star}$ indicates equal contribution.}
\blfootnote{$^{\dagger}$ State Key Laboratory of Intelligent Technology and Systems, Beijing National Research Center for Information Science and Technology.}
\setcounter{footnote}{0}

\begin{abstract}

Disparity estimation for binocular stereo images finds a wide range of applications. Traditional algorithms may fail on featureless regions, which could be handled by high-level clues such as semantic segments. In this paper, we suggest that appropriate incorporation of semantic cues can greatly rectify prediction in commonly-used disparity estimation frameworks. Our method conducts semantic feature embedding and regularizes semantic cues as the loss term to improve learning disparity. Our unified model {SegStereo} employs semantic features from segmentation and introduces semantic softmax loss, which helps improve the prediction accuracy of disparity maps. The semantic cues work well in both unsupervised and supervised manners. {SegStereo} achieves state-of-the-art results on KITTI Stereo benchmark and produces decent prediction on both CityScapes and FlyingThings3D datasets.


\end{abstract}
\section{Introduction}\label{sec:introduction}
Disparity estimation is a fundamental problem in computer vision. It is important in depth prediction, scene understanding, autonomous driving, to name a few. The main goal of disparity estimation is to find corresponding pixels from stereo images for inferring object distance according to the displacement between matching pixels.

Most previous methods~\cite{geiger2010efficient,brown2011discriminative,chen2015semantic,revaud2016deepmatching} used hand-crafted reliable features to represent image patches and then selected matching pairs. They either formulate the task as supervised learning~\cite{kong2004method,hirschmuller2009evaluation} based on current labeled dataset~\cite{Geiger2012CVPR,yamaguchi2014efficient}, or resort to unsupervised learning to form photometric loss for disparity prediction~\cite{garg2016unsupervised,monodepth17}. Recently, with the development of deep neural networks, the performance of disparity estimation is significantly improved~\cite{zbontar2016stereo}. The deep feature extracted from networks can exploit inherent global information in paired input compared to traditional methods, therefore benefits from a large number of training data either in supervised or unsupervised manner.

\begin{figure}[t]
  \centering
  \resizebox{0.98\linewidth}{!}{
   \includegraphics{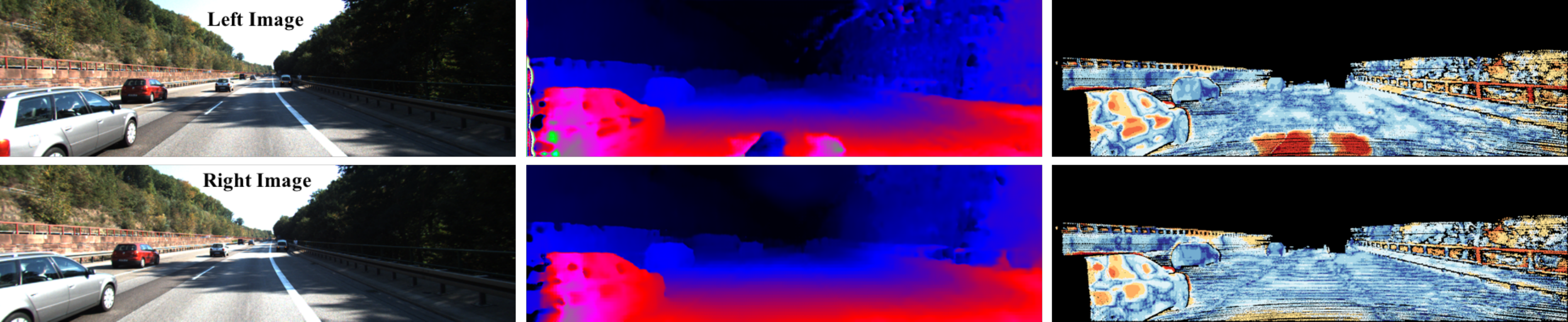} }
  \caption{Examples of prediction of unsupervised models on KITTI Stereo dataset. Left: input stereo images. Top-middle and top-right: colorized disparity and error maps predicted without semantic clues. Bottom-middle and bottom-right: colorized disparity and error maps predicted by \emph{SegStereo}. With the guidance of semantic cues, disparity estimation of \emph{SegStereo} is more accurate especially on the local ambiguous areas.}
  \label{fig:overview_results}
\end{figure}

Although deep learning based methods produce impressive feature representation given its large receptive field, it is still difficult to overcome local ambiguity, which is a common problem in disparity estimation. For example, in Fig.~\ref{fig:overview_results}, the disparity prediction in the center of road and vehicle area is not correct. It is because the matching clues for disparity estimation on those ambiguous areas are not enough to guide the model to seek correct direction for convergence, which is however the central objective for both supervised and unsupervised stereo learners.

Human can perform binocular alignment well at ambiguous areas by exploiting more cues such as global perception of foreground and background, scaling relative to the known size of familiar objects, and semantic consistency for individuals. Such ambiguous areas in disparity estimation always locate in the central region given a big target. They are easy to deal with by semantic classification. 

Based on the above-mentioned observation, we design an unified model called \emph{SegStereo} that incorporates semantic cues into backbone disparity estimation network. Basically, we use the ResNet~\cite{he2016deep} with correlation operation~\cite{dosovitskiy2015flownet} as the encoder and several deconvolutional blocks as decoder to regress a full-size disparity map. The correlation operation is designed in~\cite{dosovitskiy2015flownet} to compute matching cost volumes based on pairs of feature maps. A segmentation sub-network is employed in our model to extract semantic features that are connected to the disparity branch as the \emph{semantic feature embedding}. Moreover, we propose the warped semantic consistency via \emph{semantic loss regularization}, which further enhances robustness of disparity estimation. Both semantic and disparity evaluation is fully-convolutional so that the proposed \emph{SegStereo} enables end-to-end training. 

Our \emph{SegStereo} model with semantic clues embedded benefits both unsupervised and supervised training. In the unsupervised training, both photometric consistency loss and semantic softmax loss are computed and propagated backward. Both the semantic feature embedding and semantic softmax loss can introduce beneficial constraints of semantic consistency. The results evaluated on KITTI Stereo dataset~\cite{Menze2015CVPR} demonstrate the effectiveness of our strategies. We also apply the unsupervised model to CityScapes dataset~\cite{cordts2016cityscapes}. It yields better performance than classical SGM method~\cite{hirschmuller2008stereo}. For the supervised training scheme, we adopt the supervised regression loss instead of unsupervised photometric consistency loss to train the model, which achieves state-of-the-art results on KITTI Stereo benchmark. We further apply the \emph{SegStereo} model to FlyingThings3D dataset~\cite{mayer2016large}. It reaches high accuracy with normal fine-tuning. 

Our main contribution and achievement are summarized below.
\begin{itemize}
   \item
   We propose a unified framework called \emph{SegStereo} that incorporates semantic segmentation information into disparity estimation pipeline, where semantic consistency becomes an active guidance for disparity estimation.
   \item
   The semantic feature embedding strategy and semantic guidance softmax loss help train the system in both unsupervised and supervised manner. 
   \item
   Our method achieves state-of-the-art results on KITTI Stereo datasets. The results on CityScapes and FlyingThings3D dataset also manifest the effectiveness of our method. 
\end{itemize}

\section{Related work}

\paragraph{\textbf{Supervised Stereo Matching}}
Traditional methods design local descriptors to compute local matching cost~\cite{geiger2010efficient,heise2015fast}, followed by some global optimization steps~\cite{hirschmuller2008stereo}. Zbontar and LeCun~\cite{zbontar2016stereo} are the first to use CNN for matching cost computation. Luo~\etal~\cite{luo2016efficient} designed a siamese network that extracts marginal distributions over all possible disparities for each pixel. Chen~\etal~\cite{chen2015deep} presented a multi-scale deep embedding model that fuses feature vectors learned within different scale-spaces. Shaked and Wolf~\cite{shaked2016improved} proposed a highway network architecture with a hybrid loss that conducts multi-level comparison of image patches.

Inspired by other pixel-wise labeling tasks, the fully-convolution network (FCN)~\cite{long2015fully} was used to enable end-to-end learning of disparity maps. Mayer~\etal~\cite{mayer2016large} raised {DispNet} with a correlation module to encode matching cues instead of picking corresponding pairs from stereo images. Kendall~\etal~\cite{kendall2017end} proposed the {GC-Net} framework that combines contextual information by means of 3D convolutions over a cost volume. A three-stage network of Gidaris and Komodakis \cite{gidaris2016detect} implements a pipeline to detect, replace, and refine disparity errors respectively. Pang~\etal~\cite{pang2017cascade} presented a cascade network where the second stage learned the residual between initial result and ground-truth values.

Yu~\etal~\cite{yu2018deep} designed a two-stream network for generation and selection of cost aggregation proposals respectively. Liang~\etal~\cite{Liang2018Learning} integrated disparity estimation and refinement into one network. It reaches state-of-the-art performance on KITTI benchmark~\cite{Menze2015CVPR}. Chang and Chen~\cite{chang2018pyramid} exploited context information for finding correspondence by a pyramid stereo matching network. In contrast, our method concentrates on combining semantic information to improve disparity estimation by semantic feature embedding.

\paragraph{\textbf{Unsupervised Stereo Matching}}
In recent years, a number of unsupervised learning methods based on spatial transformation were proposed for view synthesis, depth prediction, optical flow and disparity estimation. Unsupervised methods get rid of the dependence of ground-truth labels, which are always expensive to access. Flynn~\etal~\cite{Flynn_2016_CVPR} presented an image synthesis network called {DeepStereo} that learns a cost volume combined with a separate conditional color model. Xie~\etal~\cite{xie2016deep3d} designed a {Deep3D} network that minimizes pixel-wise reconstruction loss to generate right-view images. 

Garg~\etal~\cite{garg2016unsupervised} proposed an end-to-end framework to learn single-view depth by optimizing the projection errors in a calibrated stereo environment. The improved method~\cite{monodepth17} introduces a fully-differentiable structure and an extra left-right consistency check that leads to better results. A semi-supervised approach was proposed by Kuznietsov~\etal~\cite{kuznietsov2017semi} where supervised and unsupervised alignment loss are used to train the network for depth estimation. Yu~\etal~\cite{jason2016back} focused on unsupervised learning of optical flow via photometric constancy and motion smoothness. Meister~\etal~\cite{Meister2018UnFlow} defined a bidirectional census loss to train optical flow. An iterative unsupervised learning network presented by Zhou~\etal~\cite{zhou2017stereo} adopts left-right checking to pick suitable matching pairs. Compared with these unsupervised methods, our model applies warping reconstruction to both photometric image and semantic maps, along with additional semantic feature embedding, to reliably estimate disparity.

\paragraph{\textbf{Semantic-Guided Algorithms}}

Compared to disparity estimation, semantic segmentation is a high-level classification task where each pixel in the image is assigned to a class \cite{long2015fully,badrinarayanan2015segnet,chen2015semantic,zhao2017pspnet}. Several methods apply scene parsing information to other tasks. Guney and Geiger~\cite{guney2015displets} leveraged object knowledge in MRF formulation to resolve stereo ambiguity. Bai~\etal~\cite{Bai2016Exploiting} tackled instance-level segmentation and epipolar constraints to reduce the uncertainty of optical flow estimation. A cascaded classification framework of Ren~\etal~\cite{ren2017cascaded} iteratively refines semantic masks, stereo correspondence and optical flow fields. Behl~\etal\cite{behl2017bounding} integrated the instance recognition cues into a CRF-based model for scene flow estimation. 

With similar motivation to ours, Cheng~\etal~\cite{cheng2017segflow} designed an end-to-end trainable network called {SegFlow}, which enables joint learning for video object segmentation and optical flow. This model contains a segmentation branch and a flow branch whose feature maps concatenate. We differently focus on disparity estimation, where objects in the scene are captured at the same time so that stable structural information can be exploited. In addition, our \emph{SegStereo} model also propagates softmax loss back to disparity branch by warping, which makes semantic information effective in the whole training process. In addition, our model enables unsupervised learning of disparity with photometric loss and semantic-aware constraints.

\begin{figure}[th]
  \centering
  \includegraphics[width=\linewidth]{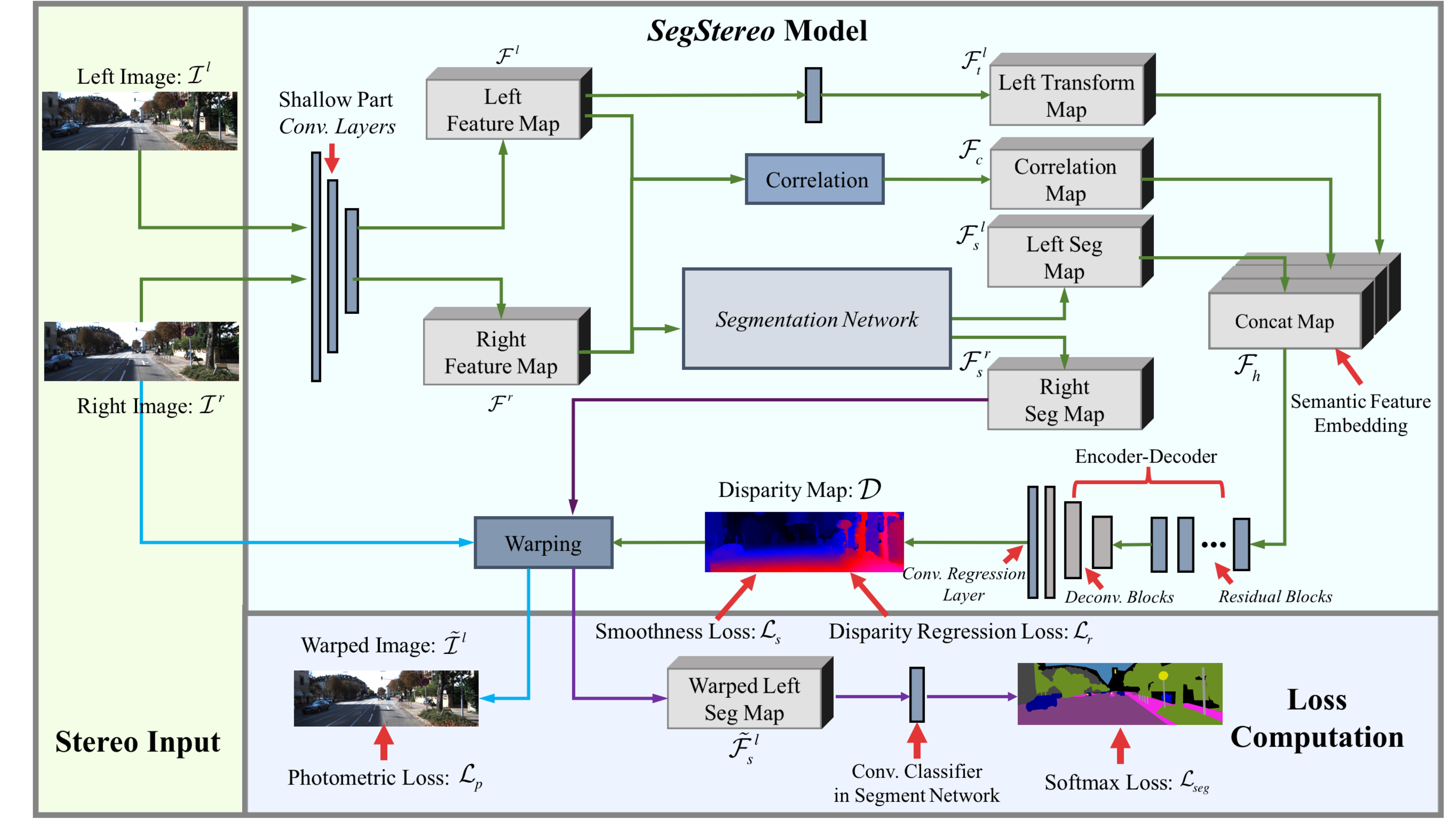}
  \caption{Our \emph{SegStereo} framework. We extract intermediate features $\mathcal{F}_{l}$ and $\mathcal{F}_{r}$ from stereo input. We calculate the cost volume $\mathcal{F}_c$ via the correlation operator. The left segmentation feature map ${\mathcal{F}_{s}}^{l}$ is aggregated into disparity branch as \emph{semantic feature embedding}. The right segmentation feature map ${\mathcal{F}_{s}}^{r}$ is warped to left view for per-pixel semantic prediction with \emph{softmax loss regularization}. Both steps incorporate semantic information to improve disparity estimation. The \emph{SegStereo} framework enables both unsupervised and supervised learning, using photometric loss $\mathcal{L}_p$ or disparity regression loss $\mathcal{L}_r$.
  }
  \label{fig:segstereo}
\end{figure}

\section{Our Method}
In this section, we describe our \emph{SegStereo} disparity estimation architecture, suitable for both unsupervised and supervised learning. We first present the basic network for disparity regression. Then we detail our incorporation strategies of semantic cues, including semantic feature embedding and semantic loss regularization. Both of them are effective to rectify disparity prediction. Finally, we show how disparity estimation is achieved under unsupervised and supervised conditions.

\subsection{Basic Network Architecture}
Our overall \emph{SegStereo} network is shown in Fig.~\ref{fig:segstereo}. The backbone network is ResNet-50~\cite{he2016deep}. Instead of directly computing disparity on raw pixels, we adopt the shallow part of ResNet-50 model to extract image features $\mathcal{F}^{l}$ and $\mathcal{F}^{r}$ on the paired input $\mathcal{I}^{l}$ and $\mathcal{I}^{r}$, which is known as robust to local context information encoding. 

The cost volume features for stereo matching $\mathcal{F}_{c}$ are computed by correlation layer between $\mathcal{F}^{l}$ and $\mathcal{F}^{r}$, similar to that of DispNetC~\cite{mayer2016large}. To preserve detail information on left stereo feature, we apply a convolution block on $\mathcal{F}^{l}$ and obtain transformed feature ${\mathcal{F}_{t}}^{l}$. Meanwhile, a segmentation network is utilized to compute semantic features ${\mathcal{F}_{s}}^{l}$ and ${\mathcal{F}_{s}}^{r}$ for left and right images respectively, sharing shallow layer representation with disparity network. The left transformed disparity features ${\mathcal{F}_{t}}^{l}$, the correlated features $\mathcal{F}_{c}$ and the left semantic features ${\mathcal{F}_{s}}^{l}$ are concatenated as hybrid feature representation $\mathcal{F}_{h}$. Here, semantic cues are preliminarily introduced to the disparity network as \emph{Semantic Feature Embedding}.

After feature embedding, we feed $\mathcal{F}_{h}$ into the disparity encoder-decoder to get full-size disparity map $\mathcal{D}$. The disparity map is further used to warp right semantic feature ${\mathcal{F}_{s}}^{r}$ to left under \emph{Semantic Loss Regularization}, detailed in Section~\ref{subsec:method_loss_regularization}. They constitute the key components of our framework. We describe more setting details in Section~\ref{subsec:exp_model_spec} and in supplementary material.

\subsection{Semantic Feature Embedding}\label{subsec:method_feature_embedding}

The basic disparity estimation frameworks work well on image patches with edges and corners where clear matching cues are located. It can be optimized with photometric loss in an unsupervised system or guided by supervised $\ell_1$ norm regularization otherwise. The remaining major issue is on flat regions, as shown in the first row of Fig.~\ref{fig:overview_results}. We use semantic cues to help prediction and rectify the final disparity map. As a result, we first incorporate the cues by embedding of semantic feature.

Our semantic feature embedding combines information from left disparity features ${\mathcal{F}_{t}}^{l}$, the correlated features $\mathcal{F}_{c}$ and the left semantic features ${\mathcal{F}_{s}}^{l}$. It contains the following advantages. (1) The employed segmentation branch shares the shallow computation with backbone disparity network for efficient computation and effective representation. (2) The semantic feature ${\mathcal{F}_{s}}^{l}$ gives more consistent representations on those flat regions compared to the disparity feature ${\mathcal{F}_{t}}^{l}$, which introduce object-level prior knowledge. (3) The low-level features and high-level recognition information are fused explicitly via the aggregation of ${\mathcal{F}_{t}}^{l}$, $\mathcal{F}_{c}$ and ${\mathcal{F}_{s}}^{l}$. The experiments in Sections~\ref{subsec:exp_unsup} and~\ref{subsec:exp_sup} further manifest that our semantic embedding helps disparity branch predict more convincing results in both unsupervised and supervised learning. In addition, the right semantic features ${\mathcal{F}_{s}}^{r}$ are reserved for the following semantic feature warping and loss regularization. 

\subsection{Semantic Loss Regularization}\label{subsec:method_loss_regularization}

The semantic information cues can also guide learning of disparity as a loss term. As shown in Fig.~\ref{fig:segstereo}, based on the predictive disparity map $\mathcal{D}$, we employ feature warping on the right segmentation map ${\mathcal{F}_{s}}^{r}$ to get the reconstructed left segmentation map $\tilde{{{\mathcal{F}}_{s}}^{l}}$, and use left segmentation ground truth labels as guidance to learn a per-pixel classifier. Finally, the semantic cues guidance loss $\mathcal{L}_{seg}$ is measured between classified warped maps and ground-truth labels. 

When training the disparity network, the semantic loss $\mathcal{L}_{seg}$ is propagated back to disparity branch through semantic convolutional classifier and feature warping layer. Along with basic photometric loss $\mathcal{L}_{p}$ or regression loss $\mathcal{L}_{r}$, semantic loss $\mathcal{L}_{seg}$ imposes extra object-aware constraints to guide disparity training. The experiments prove that semantic loss regularization can effectively resolve the local disparity ambiguities, especially in the unsupervised learning period.

\subsection{Objective Function}
The semantic information detailed above can be used in both unsupervised and supervised systems. Here we detail the loss functions in these two conditions.

\paragraph{\textbf{Unsupervised Manner}}
One image in a stereo pair can be reconstructed from the other with estimated disparity, which should be close to the original raw input. We utilize this property as photometric consistency to help learn the disparity in an unsupervised manner. Given estimated disparity $\mathcal{D}$, we apply image warping $\varPhi$ on the right image $\mathcal{I}^{r}$ and get the warped left image reconstruction as $\tilde{\mathcal{I}}^l$. Then we adopt $\ell_1$ norm to regularize the photometric consistency with photometric loss $\mathcal{L}_p$ expressed as
\begin{equation}\label{eq:photometric}
\mathcal{L}_p = \frac{1}{N}\sum_{i,j}\delta^p_{i,j}\|\tilde{\mathcal{I}}^{l}_{i,j} - \mathcal{I}^{l}_{i,j}\|_1,
\end{equation}
where $N$ is the number of pixels. $\delta^p_{i,j}$ is a mask indicator to avoid outlier as image boarder or occluded regions, where no pixel correspondence exists. If the resulting photometric difference on position $(i,j)$ is greater than a threshold $\epsilon$, $\delta^p_{i,j}$ is $0$, otherwise, it is $1$.

The photometric consistency enables disparity learning in an unsupervised manner. If there is no regularization term in $\mathcal{L}_p$ to enforce local smoothness of the estimated disparity, local disparity may be incoherent. To remedy this issue, we apply $\ell_1$ penalty to disparity gradients $\partial{\mathcal{D}}$ with the smoothness loss $\mathcal{L}_s$ defined as
\begin{equation}\label{eq:smooth}
\mathcal{L}_s = \frac{1}{N}\sum_{i,j}[\rho_s(\mathcal{D}_{i,j} - \mathcal{D}_{i+1,j}) + \rho_s(\mathcal{D}_{i,j} - \mathcal{D}_{i,j+1})],
\end{equation}
where $\rho_s(\cdot)$ is the spatial smoothness penalty implemented as generalized Charbonnier function~\cite{barron2017more}. 

With the semantic feature embedding and semantic loss, the overall loss in our unsupervised system is $\mathcal{L}_{unsup}$, containing the photometric loss $\mathcal{L}_p$, smoothness loss $\mathcal{L}_s$, and the semantic cues loss $\mathcal{L}_{seg}$. We note that disparity labels are not involved in loss computation so that disparity estimation is considered as an unsupervised learning process here. To balance learning of different loss branches, we introduce loss weights $\lambda_{p}$ for $\mathcal{L}_p$, $\lambda_{s}$ for $\mathcal{L}_s$, and $\lambda_{seg}$ for $\mathcal{L}_{seg}$. Thus the total loss $\mathcal{L}_{unsup}$ is expressed as
\begin{equation}
\mathcal{L}_{unsup} = \lambda_{p}\mathcal{L}_p + \lambda_{s}\mathcal{L}_s + \lambda_{seg}\mathcal{L}_{seg}.
\end{equation}

\paragraph{\textbf{Supervised Manner}}
The proposed semantic cues for disparity prediction also works in supervised training, where the ground truth disparity map $\hat{D}$ is provided. We directly adopt the $\ell_1$ norm to regularize prediction where the disparity regression loss $\mathcal{L}_r$ is
\begin{equation}\label{eq:regression}
\mathcal{L}_{r} = \frac{1}{N_\mathcal{V}}\sum_{i,j\in{\mathcal{V}}}\|\mathcal{D}_{i,j} - \hat{\mathcal{D}}_{i,j}\|_1,
\end{equation}
where $\mathcal{V}$ is the set of valid disparity pixels in $\hat{\mathcal{D}}$ and $N_\mathcal{V}$ is the number of valid pixels. For utilizing the semantic cues, both feature embedding and semantic softmax loss are adopted as described in Sections~\ref{subsec:method_feature_embedding} and~\ref{subsec:method_loss_regularization}. Loss weight $\lambda_{r}$ is used for regression term $\mathcal{L}_r$. The overall loss function $\mathcal{L}_{sup}$ becomes
\begin{equation}
\mathcal{L}_{sup} = \lambda_{r}\mathcal{L}_r + \lambda_{s}\mathcal{L}_s + \lambda_{seg}\mathcal{L}_{seg}.
\end{equation}

\section{Experimental Results}
In this section, we evaluate key components in the \emph{SegStereo} model. We mainly pretrain the model on CityScapes dataset~\cite{cordts2016cityscapes} and evaluate it on KITTI Stereo 2015 dataset~\cite{Menze2015CVPR}. We also compare the performance of our method with other disparity estimation methods on KITTI benchmark~\cite{Menze2015CVPR}. Further, we apply our \emph{SegStereo} model to FlyingThings3D dataset~\cite{mayer2016large} to assess performance on different scenes.

\subsection{Model Specification}\label{subsec:exp_model_spec}

PSPNet-50~\cite{zhao2017pspnet} is employed as a segmentation network due to its high performance. The layers (from ``conv1\_1'' to ``conv3\_1) of PSPNet-50 are used as the shallow part. The extracted features $\mathcal{F}_{l}$ and $\mathcal{F}_{r}$ have a $1/8$ spatial size to raw images. We select the output of ``conv5\_4'' layer of PSPNet-50 as semantic features. The weights in the shallow part and segmentation network are fixed when training \emph{SegStereo}.

For cost volume computation, we perform 1D-correlation~\cite{mayer2016large} between $\mathcal{F}^{l}$ and $\mathcal{F}^{r}$ according to epipolar constraints. Both max displacement and padding size are set to $24$ so that the channel number of correlated features $\mathcal{F}_{c}$ is $25$. For left feature transformation, the kernel size of transformed convolutional layer is $1 \times 1 \times 256$. All of $\mathcal{F}_{c}$, ${\mathcal{F}_{t}}^{l}$ and ${\mathcal{F}_{s}}^{l}$ have the same spatial size. We directly concatenate them to form the hybrid feature map $\mathcal{F}_{h}$.

Disparity encoder behind hybrid features $\mathcal{F}_{h}$ contains $12$ residual blocks. Several common convolutional operations in residual blocks are replaced with dilation patterns~\cite{zhao2017pspnet} to integrate wider context information. Disparity decoder consists of $3$ deconvolutional blocks and $1$ convolutional regression layer to output full-size disparity map. We provide more details in supplementary material. 

The right segmentation map ${\mathcal{F}_{s}}^{r}$ is of $1/8$ size to the raw image, while the estimated disparity map $D$ is of full size. To perform feature warping, we first upsample ${\mathcal{F}_{s}}^{r}$ to the full size. We afterwards downsample warped feature map to $1/8$ size and get the final reconstructed left segmentation feature map as $\tilde{{{\mathcal{F}}_{s}}^{l}}$.

\subsection{Baseline Model Excluding Semantic Information}
To validate the effect of incorporating semantic cues, we design a baseline model called \emph{ResNetCorr} without any semantic information. The hybrid features $\mathcal{F}_{h}$ in \emph{ResNetCorr} is concatenated with the correlated features $\mathcal{F}_{c}$ and left transformed features ${\mathcal{F}_t}^{l}$. The rest encoder-decoders are attached behind $\mathcal{F}_{h}$, as that of \emph{SegStereo}. The softmax $L_{seg}$ term is excluded in loss computation. We provide the structural definition of \emph{ResNetCorr} model in supplementary material.

\subsection{Datasets and Evaluation Metrics}

The CityScapes dataset~\cite{cordts2016cityscapes} is released for urban scene understanding. It provides rectified stereo image pairs and corresponding disparity maps pre-computed by SGM algorithm~\cite{hirschmuller2008stereo}. It contains 5,000 high quality pixel-level finely annotated maps for left-view. These images are split into sets with numbers $2,975$, $500$ and $1,525$ for training, validation and testing. In addition, this dataset provides $19,997$ stereo images and their SGM labels in extra training set. We will use these extra data for model pretraining. 

The KITTI Stereo 2015 dataset~\cite{Menze2015CVPR} contains 200 training and 200 testing image pairs. The 200 training images also has semantic labels~\cite{Alhaija2017BMVC}. We mainly use the dataset for fine tuning and evaluation. The KITTI Stereo 2012 dataset~\cite{Geiger2012CVPR} also provides disparity maps, which contain 194 training and 195 testing image pairs. 

The FlyingThings3D dataset~\cite{mayer2016large} is a virtual dataset for scene matching including optical flow estimation and disparity prediction. This dataset is rendered by computer graphics techniques with background objects and 3D models. It provides 22,390 images for training and 4,370 images for testing. 

To evaluate the results, we apply the end-point-error (EPE), which measures the average pixel deviation and the bad pixel error (D1). The latter calculates the percentage of disparity errors below a threshold. Both the errors in non-occluded region (Noc) and all pixels (All) are evaluated.

\subsection{Implementation Details}

Our implementation of the \emph{SegStereo} model is based on a customized Caffe~\cite{jia2014caffe}. We use the ``poly'' learning rate policy where current learning rate equals to the base one multiplying $(1-\frac{iter}{max\_iter})^{power}$. Such learning policy is also adopted in~\cite{brown2011discriminative,zhao2017pspnet} for better performance. When training on CityScapes dataset, we set base learning rate to $0.01$, power to $0.9$. Momentum and weight decay are set to $0.9$ and $0.0001$, respectively. These parameters of learning policy are kept on supervised fine-tuning process. 

For data augmentation, we adopt random resizing, color shift and contrast brightness adjustment. The random factor is between $0.5$ to $2.0$. The maximum color shift along RGB axes is set to $10$ and the maximum brightness shift is set to $5$. The contrast multiplier is between $0.8$ and $1.2$. The ``cropsize'' is set to $513 \times 513$ and batch size is set to $16$. 

In unsupervised training, the loss weights $\lambda_{p}$, $\lambda_{s}$ and $\lambda_{seg}$ for photometric, softmax and smoothness terms are set to $1.0$, $10.0$, $0.1$, respectively. The threshold $\epsilon$ in photometric loss is set to $10$. When switching to supervised training, if providing semantic labels, the loss weights for $\lambda_{r}$, $\lambda_{s}$ and $\lambda_{seg}$ for regression, softmax and smoothness term are set to $1.0$, $1.0$, $0.1$. If no semantic labels are provided, the loss weight of softmax term is set to $0$. The Charbonnier parameters $\alpha$, $\beta$ and $\epsilon$ in smoothness loss term are $0.21$, $5.0$ and $0.001$ as described in~\cite{jason2016back}. 

\begin{table}[!t]
  \centering
  \caption{Results of unsupervised training models on KITTI Stereo 2015~\cite{Menze2015CVPR}.}
  \resizebox{0.90\textwidth}{!}{\input{unsuper_results.tex}}
  \label{tab:unsupervised_results}   
\end{table}

\subsection{Unsupervised Learning}\label{subsec:exp_unsup}

\paragraph{\textbf{Semantic Feature Embedding}} The first experiment in Tab.~\ref{tab:unsupervised_results} compares the errors between \emph{ResNetCorr} and \emph{SegStereo} models. We observe that with semantic features from PSPNet-50, \emph{EPE} is improved by 20\% and the \emph{D1 error} is reduced by 15\% when only adopting photometric loss. When combining the photometric and smoothness losses to train the models, \emph{EPE} is improved by 12\% and the \emph{D1 error} is reduced by 13\%. It shows that semantic feature embedding significantly reduces the disparity errors.

\paragraph{\textbf{Softmax Loss Regularization}} The second experiment in Tab.~\ref{tab:unsupervised_results} is to validate the effect of softmax loss regularization. Based on photometric loss, we use smoothness loss to penalize discontinuity on disparity maps, which reduces \emph{EPE} from 2.72 to 2.17 and the D1 error from 12.08 to 10.53 on all pixels. With additional softmax loss to constrain semantic consistency, \emph{EPE} decreases from 2.17 to 1.89 and the \emph{D1 error} decreases from 10.53 to 10.03. Thus, the regularization of softmax loss reduces \emph{EPE} by 13\% and the \emph{D1 error} by 5\%, respectively. 

\begin{figure}[!t]
  \centering
  \includegraphics[width=\textwidth]{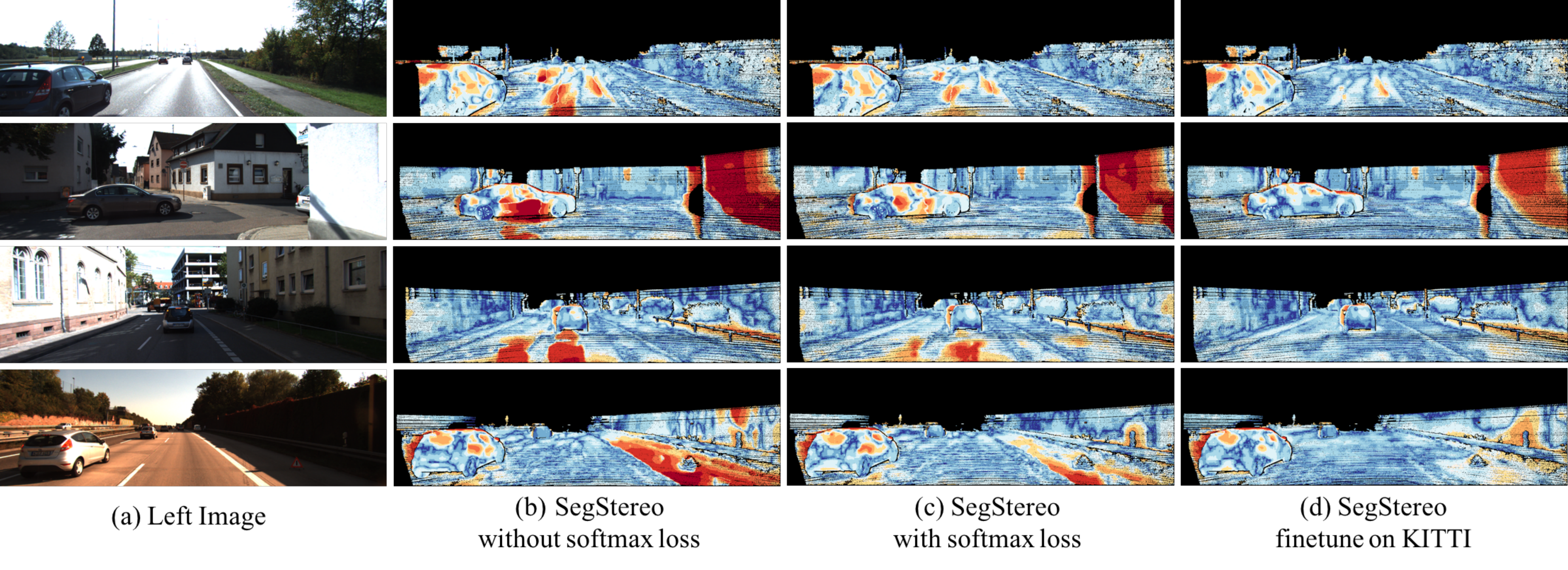}
  \caption{Qualitative examples of unsupervised \emph{SegStereo} models on KITTI Stereo 2015 dataset~\cite{Menze2015CVPR}. With the guidance of softmax regularization and additional fine-tune process, the accuracy of disparity is improved.}
  \label{fig:unsupervised_qualitative_results}   
\end{figure}

\begin{figure}[!t]
  \centering
  \includegraphics[width=0.85\textwidth]{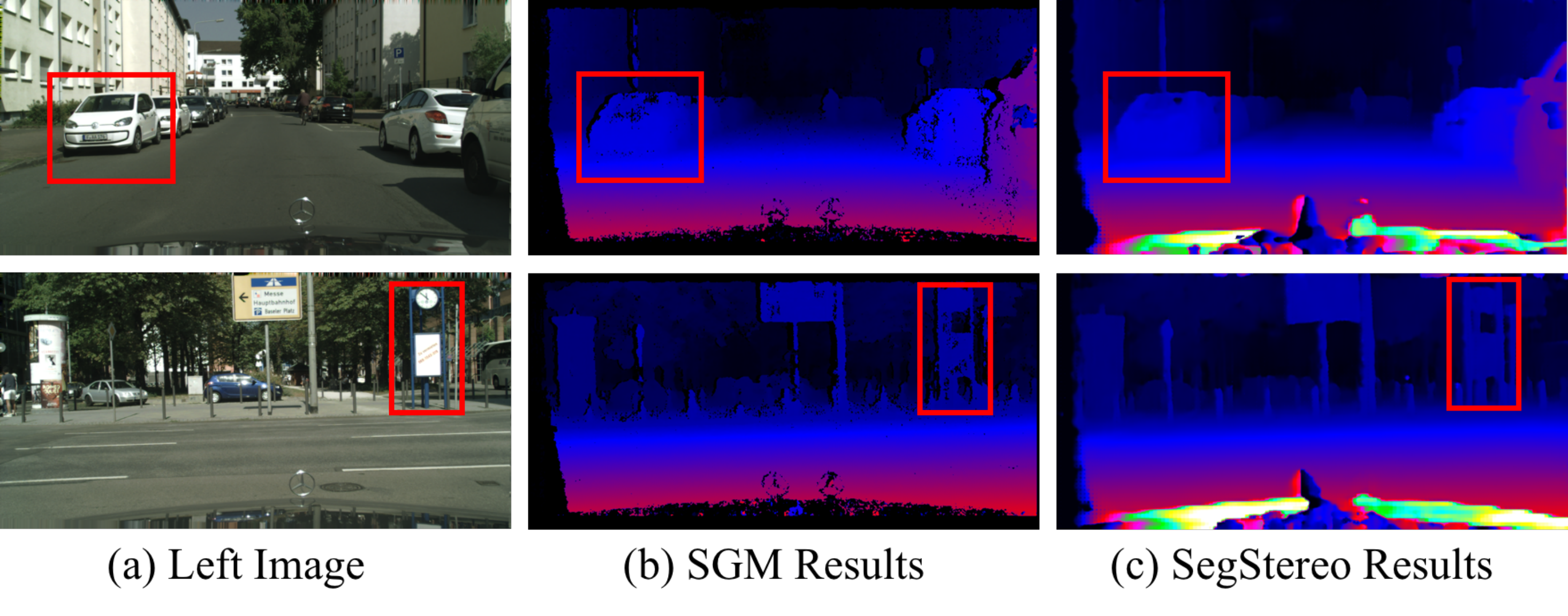}
  \caption{Qualitative examples of unsupervised-learning version of the \emph{SegStereo} model on CityScapes validation set~\cite{cordts2016cityscapes}. From left to right: left input images, disparity maps predicted by SGM algorithm \cite{hirschmuller2008stereo}, and our disparity maps.}
  \label{fig:unsupervised_cityscapes_examples}
\end{figure}

Fig. \ref{fig:unsupervised_qualitative_results} shows results of different loss combinations (with or without softmax loss). We observe that the gain of softmax loss mainly arises on big objects, such as road and car, which directly help enhance disparity prediction on local ambiguous regions.

\paragraph{\textbf{Finetune on KITTI Stereo dataset}}
We compare our approach with other unsupervised methods in the third experiment as listed in Tab.~\ref{tab:unsupervised_results}. To adapt our model to KITTI dataset, we finely tune \emph{SegStereo} on the 200 images of KITTI 2015 training set. We set the maximum iteration number to 500 and batch size to 16, so that 40 epochs are conducted. All photometric loss, smoothness loss and softmax loss are used in this process. Qualitative results in Fig.~\ref{fig:unsupervised_qualitative_results} show that prediction errors are further reduced by fine-tuning. Our model outperforms the other two unsupervised methods~\cite{zhou2017stereo,monodepth17} on KITTI 2015 benchmark. 

\paragraph{\textbf{CityScapes Results}}  We adapt the unsupervised \emph{SegStereo} model to CityScapes dataset~\cite{cordts2016cityscapes}. In Fig.~\ref{fig:unsupervised_cityscapes_examples}, we give several examples to visualize quality on the validation set. Compared to the results of SGM algorithm~\cite{hirschmuller2008stereo}, our method yields better structures in term of global scene information and details of objects. 

\begin{table}[!t]
  \centering
  \caption{Results of supervised-training models evaluated on KITTI Stereo 2015~\cite{Menze2015CVPR}}
  \resizebox{0.90\textwidth}{!}{\input{super_results.tex}}
  \label{tab:supervised_results}  
\end{table}

\subsection{Supervised Learning}\label{subsec:exp_sup}

\paragraph{\textbf{KITTI Results }} In supervised learning, the ground-truth disparity maps are directly applied to train our \emph{SegStereo} model. As KITTI stereo dataset is too small, we pre-train our model on CityScapes dataset. Although the disparity maps computed by SGM algorithm contain errors and holes, they are useful for our model to get reasonable accuracy. The maximum iteration is set to $90K$. Different from unsupervised training, here the disparity regression loss $\mathcal{L}_r$ plays the major role. We also compare the performance between \emph{ResNetCorr} and \emph{SegStereo}. The first experiment in Tab.~\ref{tab:supervised_results} shows that disparity error rate is slightly reduced by semantic feature embedding when we pretrain the models on CityScapes dataset~\cite{cordts2016cityscapes}. 

In the second experiment, we fuse the extra training set in CityScapes and training set in FlyingThings3D dataset to pretrain \emph{ResNetCorr} and \emph{SegStereo}. Since there is no semantic labels in such two datasets, we do not compute softmax loss. The weights in segmentation branch of \emph{SegStereo} is pretrained on CityScapes training set and fixed. We extend the maximum iterations to $500K$. Compared to the first experiment, with more training data, both \emph{ResNetCorr} and \emph{SegStereo} achieve higher accuracy. And the performance of \emph{SegStereo} is still better than \emph{ResNetCorr}.

\begin{table}[h]
  \centering
  \caption{Comparison with other disparity estimation methods on the test set of KITTI 2015 \cite{Menze2015CVPR}. Our method achieves state-of-the-art results on this benchmark.}
  \resizebox{0.90\textwidth}{!}{\input{kitti_2015_test_results.tex}}
  \label{tab:kitti_test_result}
\end{table}

In the third experiment, we use KITTI Stereo 2012 and 2015 datasets to finely tune our pretrained models from the second experiment. We set the maximum iteration to 90K and base learning rate to 0.01. To facilitate performance comparison, we split Stereo 2015 training set \cite{luo2016efficient} so that 40 images are randomly selected for validation and the remaining 160 images are used for training. Tab.~\ref{tab:supervised_results} lists errors on both training and validation sets. Compared to the \emph{ResNetCorr} model, semantic feature embedding prevents overfitting and brings a certain improvement on disparity estimates. 

To exploit more detailed matching cues on fine scales, we redesign \emph{SegStereo}, where the shallow part is end with the ``conv1\_3'' layer of PSPNet-50. To adapt to the increased feature map size, the maximum displacement and padding size of the correlation layer are both set to $96$. We also up-sample the semantic feature maps from ``conv5\_4'' layer for semantic feature embedding. This redesigned model is also pretrained on the fusion set of CityScapes and FlyingThings3D, followed by fine-tuning on KITTI Stereo dataset. The new \emph{SegStereo} model (with remark ``corr13'' in Tab.~\ref{tab:supervised_results}) outperforms general \emph{SegStereo} by leveraging more detail information.

\begin{figure}[!t]
  \centering
  \includegraphics[width=0.90\linewidth]{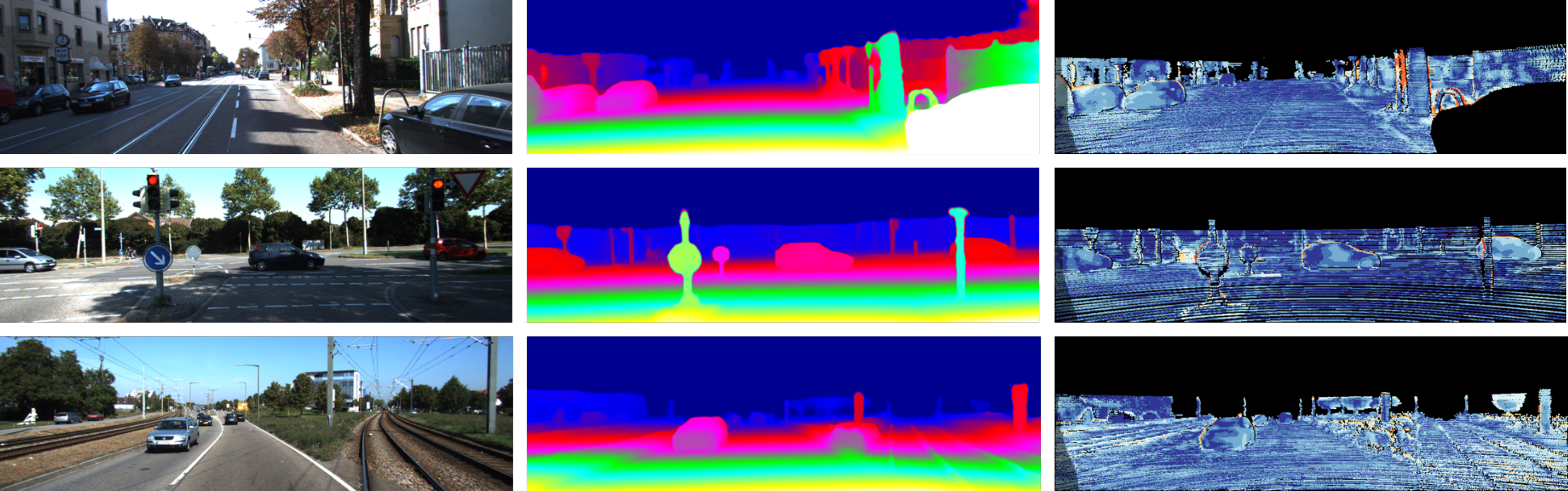}
  \caption{Supervised-learning results on KITTI Stereo 2015 test sets~\cite{Menze2015CVPR}. By incorporating semantic information, our method is able to estimate accurate disparity. From left to right, we show left input images, disparity predictions of \emph{SegStereo}, and error maps}  
  \label{supervised_qualitative_results}
\end{figure}

Tab.~\ref{tab:kitti_test_result} compares our model to other approaches on KITTI 2015 benchmark~\cite{Menze2015CVPR}. Our method achieves state-of-the-art results. Fig.~\ref{supervised_qualitative_results} gives several visual examples on KITTI 2015 test set. By incorporating semantic information, our \emph{SegStereo} model is able to handle challenging scenarios. In supplementary material, we also provide results on KITTI 2012 benchmark~\cite{Geiger2012CVPR} and segmentation results.

\begin{figure}[t!]
  \centering
  \resizebox{0.80\linewidth}{!}{
    \includegraphics{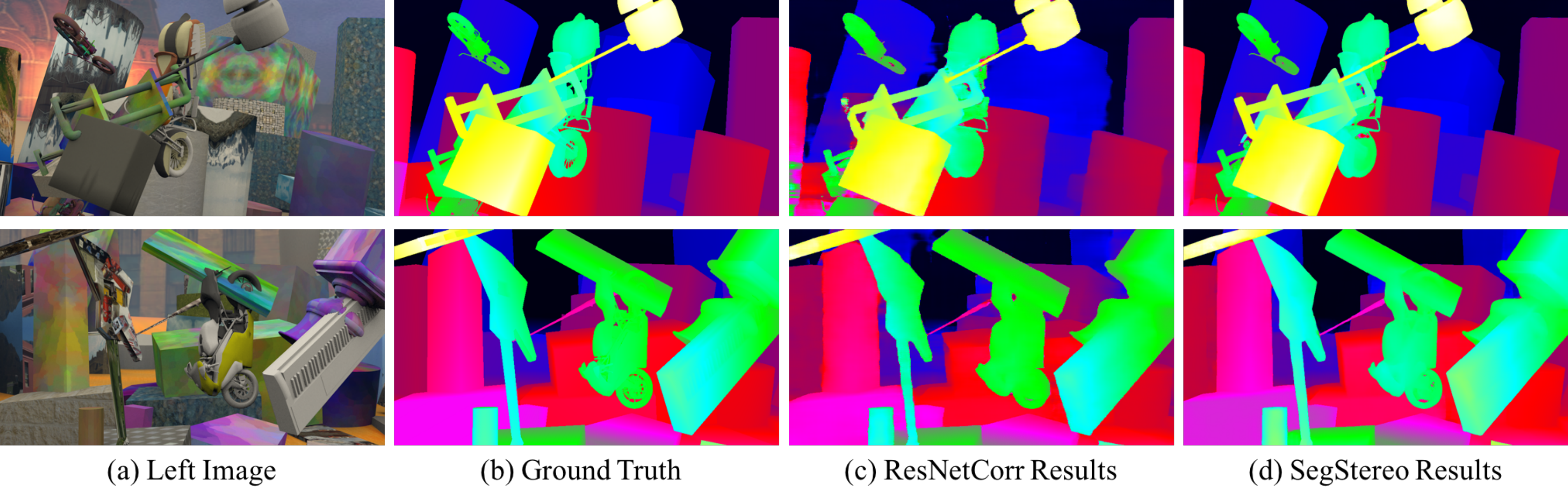}}
  \caption{Qualitative examples of \emph{ResNetCorr} and \emph{SegStereo} model on FlyingThings3D validation set~\cite{mayer2016large}. From left to right, left images, ground-truth, \emph{ResNetCorr} results and \emph{SegStereo} results}
  \label{fig:fly_examples}  
\end{figure}

\begin{table}[t]
  \centering
  \caption{Comparison with other disparity estimation methods on the test set of FlyingThings3D~\cite{mayer2016large}.}
  \resizebox{\textwidth}{!}{\input{fly_results.tex}}
  \label{tab:fly_results}
\end{table}

\paragraph{\textbf{FlyingThings3D Results }} To illustrate that our \emph{SegStereo} model can adapt to other datasets, we test the supervised-training \emph{ResNetCorr} and \emph{SegStereo} on FlyingThings3D dataset~\cite{mayer2016large}. Here, we directly select the pretrained models from the second experiments of Tab.~\ref{tab:supervised_results}. The two models are compared with other methods on the validation set of FlyingThings3D in Tab.~\ref{tab:fly_results}. With the guidance of semantic information, the \emph{SegStereo} model outperforms \emph{ResNetCorr} and becomes state-of-the-art, which indicates that segmentation modules is effective and general for disparity estimation across various datasets. Fig.~\ref{fig:fly_examples} shows several visual examples on validation set. 

\section{Conclusion}

In this paper, we have proposed a unified model \emph{SegStereo}, which integrates semantic feature maps into disparity prediction pipeline. A softmax loss is combined with common photometric loss or disparity regression loss to enable training in both unsupervised and supervised manners. Our \emph{SegStereo} leads to more reliable results, especially on ambiguous areas. Experiments on KITTI Stereo datasets demonstrate the effectiveness of the semantic-guided strategy. Our method achieves state-of-the-art performance on this benchmark. Results on CityScapes and FlyingThings3D datasets further manifest its adaptability.

\section*{Acknowledgment}

This work was supported in part by the National Key R\&D Program of China under Grant No. 2017YFB1302200 and by Joint Fund of NORINCO Group of China for Advanced Research under Grant No. 6141B010318.

\clearpage

\bibliographystyle{splncs04}
\bibliography{egbib}

\end{document}

%% file: unsuper_results.tex
\begin{tabular}{l | c c | c c }
    \toprule[1pt]
    \multirow{2}{*}{Model} &
    \multicolumn{2}{c|}{Noc pixels} &
    \multicolumn{2}{c}{All pixels} \\
    & {~~~EPE~~~} & {~D1 Error~} & {~~~EPE~~~} & {~D1 Error~} \\
    \hline
    \hline
    \multicolumn{5}{c}{\emph{1. Evaluation of semantic feature embedding}} \\
    \hline
    ResCorr (photometric loss)                                    & 2.46 & 12.78 & 3.36 & 14.08 \\
    \textbf{SegStereo (photometric loss)}                         & \textbf{1.98} & \textbf{10.76} & \textbf{2.72} & \textbf{12.08} \\  
    ResCorr (photometric loss + smooth loss)   & 2.13 & 11.05 & 2.43 & 12.16 \\
    \textbf{SegStereo (photometric loss + smooth loss)} & \textbf{1.87} & \textbf{9.39} & \textbf{2.17} & \textbf{10.53} \\
    \hline
    \multicolumn{5}{c}{\emph{2. Evaluation of softmax loss regularization}} \\
    \hline
    SegStereo  (photometric)                        & 1.98 & 10.76 & 2.72 & 12.08 \\  
    SegStereo  (photometric + smooth)               & 1.87 &  9.39 & 2.17 & 10.53 \\
    \textbf{SegStereo  (photometric + smooth + softmax)}     & \textbf{1.61} &  \textbf{8.95} & \textbf{1.89} & \textbf{10.03} \\
    \hline
    \multicolumn{5}{c}{\emph{3. Comparison to other unsupervised methods}} \\
    \hline
    Zhou \cite{zhou2017stereo}                      &  --  &  8.61 &  --  &  9.91 \\
    Godard \cite{monodepth17}                       &  --  &  --   &  --  &  9.19 \\
    SegStereo (pretrain on Cityscapes dataset)      & 1.61 &  8.95 & 1.89 & 10.03 \\
    \textbf{SegStereo (ft on KITTI Stereo dataset)}    & \textbf{1.46} &  \textbf{7.70} & \textbf{1.84} & \textbf{8.79} \\ 
    \bottomrule[1pt]
\end{tabular}

%% file: super_results.tex
\begin{tabular}{ l | c c | c c | c c | c c }
    \toprule[1pt]
    \multirow{4}{*}{Model}  & \multicolumn{4}{|c|}{Train}  & \multicolumn{4}{|c}{Test}    \\ 
    & \multicolumn{2}{|c|}{EPE}    & \multicolumn{2}{|c|}{D1} 
    & \multicolumn{2}{|c|}{EPE}    & \multicolumn{2}{|c}{D1}       \\
    {}                      & {~~~Noc~~~} & {~~~All~~~} & {~~~Noc~~~} & {~~~All~~~} 
                            & {~~~Noc~~~} & {~~~All~~~} & {~~~Noc~~~} & {~~~All~~~} \\
    \hline
    \hline
    \multicolumn{9}{c}{\emph{1. Pretrained on Cityscapes dataset}} \\
    \hline
    ResNetCorr             & -- & -- & -- & -- & 1.43 & 1.46 & 7.33 & 7.64 \\
    \textbf{SegStereo}     & -- & -- & -- & -- & \textbf{1.39} & \textbf{1.41} & \textbf{7.01} & \textbf{7.34} \\
    \hline
    \multicolumn{9}{c}{\emph{2. Pretrained on Cityscapes extra set and FlyingThings3D dataset}} \\
    \hline
    ResNetCorr             & -- & -- & -- & -- & 1.19 & 1.21 & 5.46 & 5.64 \\
    \textbf{SegStereo}     & -- & -- & -- & -- & \textbf{1.15} & \textbf{1.17} & \textbf{5.20} & \textbf{5.38} \\
    \hline
    \multicolumn{9}{c}{\emph{3. Finetune on KITTI stereo 2012 and 2015 dataset}} \\
    \hline
    ResNetCorr             & 0.40  & 0.41  & 0.68  & 0.76  & 0.73 & 0.76 & 2.13 & 2.40 \\
    \textbf{SegStereo}     & 0.40  & 0.41  & 0.65  & 0.70  & 0.73 & 0.75 & 2.11 & 2.30 \\
    \textbf{SegStereo (corr13)}    & 0.39 & 0.40 & 0.65 & 0.70 & \textbf{0.66} & \textbf{0.70} & \textbf{1.96} & \textbf{2.25}  \\
    \bottomrule[1pt]
\end{tabular}

%% file: kitti_2015_test_results.tex
\begin{tabular}{ l | c c c | c  c  c | c}
      \toprule[1pt]
      \multirow{2}{*}{Methods} & \multicolumn{3}{|c|}{Noc} & \multicolumn{3}{|c|}{All} & Runtime \\
      {} & {~~D1-bg~~} & {~~D1-fg~~} & {~~D1-all~~} & {~~D1-bg~~} & {~~D1-fg~~} & {~~D1-all~~} & {~(s)~} \\
      \hline\hline
      SPS-st~\cite{yamaguchi2014efficient}      & 3.50 & 11.61 & 4.84  & 3.84 & 12.67 & 5.31  & 2   \\
      Content-CNN~\cite{luo2016efficient}       & 3.32 & 7.44  & 4.00  & 3.73 & 8.58  & 4.54  & 1   \\
      DispNetC~\cite{mayer2016large}            & 4.11 & 3.72  & 4.05  & 4.32 & 4.41  & 4.34  & \textbf{0.06} \\
      MC-CNN~\cite{zbontar2016stereo}           & 2.48 & 7.64  & 3.33  & 2.89 & 8.88  & 3.89  & 67  \\
      PBCP~\cite{seki2016patch}                 & 2.27 & 7.71  & 3.17  & 2.58 & 8.74  & 3.61  & 68  \\
      Displets v2~\cite{guney2015displets}      & 2.73 & 4.95  & 3.09  & 3.00 & 5.56  & 3.43  & 265 \\
      L-ResMatch~\cite{shaked2016improved}      & 2.35 & 5.74  & 2.91  & 2.72 & 6.95  & 3.42  & 48  \\
      DRR~\cite{gidaris2016detect}              & 2.34 & 4.87  & 2.76  & 2.58 & 6.04  & 3.16  & 0.4 \\
      GC-NET~\cite{kendall2017end}              & 2.02 & 5.58  & 2.61  & 2.21 & 6.16  & 2.87  & 0.9 \\
      CRL~\cite{pang2017cascade}                & 2.32 & 3.12  & 2.45  & 2.48 & 3.59  & 2.67  & 0.47 \\ 
      DeepStereo~\cite{yu2018deep}              & 2.06 & 5.32  & 2.32  & 2.17 & 5.46  & 2.79  & 1.13 \\
      iResNet~\cite{Liang2018Learning}          & 2.07 & \textbf{2.76}  & 2.19  & 2.25 & \textbf{3.40}  & 2.44  & 0.12 \\
      PSMNet~\cite{chang2018pyramid}            & \textbf{1.71} & 4.31  & 2.14  & \textbf{1.86} & 4.62  & 2.32  & 0.41 \\
      \hline
      \textbf{SegStereo(Ours)}                  & 1.76 & 3.70   & \textbf{2.08}  
                                                & 1.88 & 4.07   & \textbf{2.25}  & 0.6  \\
      \bottomrule[1pt]
\end{tabular}

%% file: fly_results.tex
\begin{tabular}{ c | c c c c c c c }
  \toprule
  {~Model~~}  & {~SGM~\cite{hirschmuller2008stereo}~~} 
              & {~DispNetC~\cite{mayer2016large}}
              & {~GC-Net~\cite{kendall2017end}~~} 
              & {~CRL~\cite{pang2017cascade}~~}
              & {~iResNet~\cite{Liang2018Learning}~~}
              & {~\textbf{ResNetCorr}~~}
              & {~\textbf{SegStereo}~~}  \\
  \hline
  {~EPE~~}     & 7.29  & 2.33  & 1.84 & 1.67 & \textbf{1.27} & 3.50 & 1.45 \\
  {~D1~~}      & 16.18 & 10.04 & 9.67 & 6.70 & 4.90 & 8.45 & \textbf{3.50} \\
  \bottomrule
\end{tabular}

%% file: segstereo.bbl
\begin{thebibliography}{10}
\providecommand{\url}[1]{\texttt{#1}}
\providecommand{\urlprefix}{URL }
\providecommand{\doi}[1]{https://doi.org/#1}

\bibitem{Alhaija2017BMVC}
Alhaija, H.A., Mustikovela, S.K., Mescheder, L., Geiger, A., Rother, C.:
  Augmented reality meets deep learning for car instance segmentation in urban
  scenes. In: BMVC (2017)

\bibitem{Bai2016Exploiting}
Bai, M., Luo, W., Kundu, K., Urtasun, R.: Exploiting semantic information and
  deep matching for optical flow. In: ECCV (2016)

\bibitem{barron2017more}
Barron, J.T.: A more general robust loss function. arXiv preprint
  arXiv:1701.03077  (2017)

\bibitem{behl2017bounding}
Behl, A., Jafari, O.H., Mustikovela, S.K., Alhaija, H.A., Rother, C., Geiger,
  A.: Bounding boxes, segmentations and object coordinates: How important is
  recognition for 3d scene flow estimation in autonomous driving scenarios? In:
  ICCV (2017)

\bibitem{brown2011discriminative}
Brown, M., Hua, G., Winder, S.: Discriminative learning of local image
  descriptors. TPAMI  (2011)

\bibitem{chang2018pyramid}
Chang, J., Chen, Y.: Pyramid stereo matching network. In: CVPR (2018)

\bibitem{chen2015semantic}
Chen, L., Papandreou, G., Kokkinos, I., Murphy, K., Yuille, A.L.: Semantic
  image segmentation with deep convolutional nets and fully connected crfs
  (2015)

\bibitem{chen2015deep}
Chen, Z., Sun, X., Wang, L., Yu, Y., Huang, C.: A deep visual correspondence
  embedding model for stereo matching costs. In: ICCV (2015)

\bibitem{cheng2017segflow}
Cheng, J., Tsai, Y.H., Wang, S., Yang, M.H.: Segflow: Joint learning for video
  object segmentation and optical flow. In: ICCV (2017)

\bibitem{cordts2016cityscapes}
Cordts, M., Omran, M., Ramos, S., Rehfeld, T., Enzweiler, M., Benenson, R.,
  Franke, U., Roth, S., Schiele, B.: The cityscapes dataset for semantic urban
  scene understanding. In: CVPR (2016)

\bibitem{dosovitskiy2015flownet}
Dosovitskiy, A., Fischer, P., Ilg, E., Hausser, P., Hazirbas, C., Golkov, V.,
  van~der Smagt, P., Cremers, D., Brox, T.: Flownet: Learning optical flow with
  convolutional networks. In: ICCV (2015)

\bibitem{Flynn_2016_CVPR}
Flynn, J., Neulander, I., Philbin, J., Snavely, N.: Deepstereo: Learning to
  predict new views from the world's imagery. In: CVPR (2016)

\bibitem{garg2016unsupervised}
Garg, R., Carneiro, G., Reid, I.: Unsupervised cnn for single view depth
  estimation: Geometry to the rescue. In: ECCV (2016)

\bibitem{Geiger2012CVPR}
Geiger, A., Lenz, P., Urtasun, R.: Are we ready for autonomous driving? the
  kitti vision benchmark suite. In: CVPR (2012)

\bibitem{geiger2010efficient}
Geiger, A., Roser, M., Urtasun, R.: Efficient large-scale stereo matching. In:
  ACCV (2010)

\bibitem{gidaris2016detect}
Gidaris, S., Komodakis, N.: Detect, replace, refine: Deep structured prediction
  for pixel wise labeling. In: CVPR (2017)

\bibitem{monodepth17}
Godard, C., {Mac Aodha}, O., Brostow, G.J.: Unsupervised monocular depth
  estimation with left-right consistency. In: CVPR (2017)

\bibitem{guney2015displets}
Guney, F., Geiger, A.: Displets: Resolving stereo ambiguities using object
  knowledge. In: CVPR (2015)

\bibitem{he2016deep}
He, K., Zhang, X., Ren, S., Sun, J.: Deep residual learning for image
  recognition. In: CVPR (2016)

\bibitem{heise2015fast}
Heise, P., Jensen, B., Klose, S., Knoll, A.: Fast dense stereo correspondences
  by binary locality sensitive hashing. In: ICRA (2015)

\bibitem{hirschmuller2008stereo}
Hirschmuller, H.: Stereo processing by semiglobal matching and mutual
  information. TPAMI  (2008)

\bibitem{hirschmuller2009evaluation}
Hirschmuller, H., Scharstein, D.: Evaluation of stereo matching costs on images
  with radiometric differences. TPAMI  (2009)

\bibitem{jason2016back}
Jason, J.Y., Harley, A.W., Derpanis, K.G.: Back to basics: Unsupervised
  learning of optical flow via brightness constancy and motion smoothness. In:
  ECCV Workshop (2016)

\bibitem{jia2014caffe}
Jia, Y., Shelhamer, E., Donahue, J., Karayev, S., Long, J., Girshick, R.B.,
  Guadarrama, S., Darrell, T.: Caffe: Convolutional architecture for fast
  feature embedding. In: ACM MM (2014)

\bibitem{kendall2017end}
Kendall, A., Martirosyan, H., Dasgupta, S., Henry, P., Kennedy, R., Bachrach,
  A., Bry, A.: End-to-end learning of geometry and context for deep stereo
  regression. In: ICCV (2017)

\bibitem{kong2004method}
Kong, D., Tao, H.: A method for learning matching errors for stereo
  computation. In: BMVC (2004)

\bibitem{kuznietsov2017semi}
Kuznietsov, Y., St{\"u}ckler, J., Leibe, B.: Semi-supervised deep learning for
  monocular depth map prediction. In: CVPR (2017)

\bibitem{Liang2018Learning}
Liang, Z., Feng, Y., Guo, Y., Liu, H., Chen, W., Qiao, L., Z., L., Z., J.:
  Learning for disparity estimation through feature constancy. In: CVPR (2018)

\bibitem{long2015fully}
Long, J., Shelhamer, E., Darrell, T.: Fully convolutional networks for semantic
  segmentation. In: CVPR (2015)

\bibitem{luo2016efficient}
Luo, W., Schwing, A.G., Urtasun, R.: Efficient deep learning for stereo
  matching. In: CVPR (2016)

\bibitem{mayer2016large}
Mayer, N., Ilg, E., Hausser, P., Fischer, P., Cremers, D., Dosovitskiy, A.,
  Brox, T.: A large dataset to train convolutional networks for disparity,
  optical flow, and scene flow estimation. In: CVPR (2016)

\bibitem{Meister2018UnFlow}
Meister, S., Hur, J., Roth, S.: Unflow: Unsupervised learning of optical flow
  with a bidirectional census loss. In: AAAI (2018)

\bibitem{Menze2015CVPR}
Menze, M., Geiger, A.: Object scene flow for autonomous vehicles. In: CVPR
  (2015)

\bibitem{pang2017cascade}
Pang, J., Sun, W., Ren, J., Yang, C., Yan, Q.: Cascade residual learning: A
  two-stage convolutional neural network for stereo matching. In: ICCV Workshop
  (2017)

\bibitem{ren2017cascaded}
Ren, Z., Sun, D., Kautz, J., Sudderth, E.B.: Cascaded scene flow prediction
  using semantic segmentation. In: ICCV Workshop (2017)

\bibitem{revaud2016deepmatching}
Revaud, J., Weinzaepfel, P., Harchaoui, Z., Schmid, C.: Deepmatching:
  Hierarchical deformable dense matching. IJCV  (2016)

\bibitem{seki2016patch}
Seki, A., Pollefeys, M.: Patch based confidence prediction for dense disparity
  map. In: BMVC (2016)

\bibitem{shaked2016improved}
Shaked, A., Wolf, L.: Improved stereo matching with constant highway networks
  and reflective confidence learning. In: CVPR (2017)

\bibitem{badrinarayanan2015segnet}
Vijay, B., Alex, K., Cipolla, R.: Segnet: A deep convolutional encoder-decoder
  architecture for image segmentation. TPAMI  (2017)

\bibitem{xie2016deep3d}
Xie, J., Girshick, R., Farhadi, A.: Deep3d: Fully automatic 2d-to-3d video
  conversion with deep convolutional neural networks. In: ECCV (2016)

\bibitem{yamaguchi2014efficient}
Yamaguchi, K., McAllester, D., Urtasun, R.: Efficient joint segmentation,
  occlusion labeling, stereo and flow estimation. In: ECCV (2014)

\bibitem{yu2018deep}
Yu, L., Wang, Y., Wu, Y., Jia, Y.: Deep stereo matching with explicit cost
  aggregation sub-architecture. In: AAAI (2018)

\bibitem{zbontar2016stereo}
Zbontar, J., LeCun, Y.: Stereo matching by training a convolutional neural
  network to compare image patches. JMLR  (2016)

\bibitem{zhao2017pspnet}
Zhao, H., Shi, J., Qi, X., Wang, X., Jia, J.: Pyramid scene parsing network.
  In: CVPR (2017)

\bibitem{zhou2017stereo}
Zhou, C., Zhang, H., Shen, X., Jia, J.: Unsupervised learning of stereo
  matching. In: ICCV (2017)

\end{thebibliography}
